# Commonsense knowledge adversarial dataset that challenges ELECTRA*

Gongqi Lin, Yuan Miao, Xiaoyong Yang, Wenwu Ou, Lizhen Cui, Wei Guo, Chunyan Miao

*Abstract—* Commonsense knowledge is critical in human reading comprehension. While machine comprehension has made significant progress in recent years, the ability in handling commonsense knowledge remains limited. Synonyms are one of the most widely used commonsense knowledge. Constructing adversarial dataset is an important approach to find weak points of machine comprehension models and support the design of solutions. To investigate machine comprehension models' ability in handling the commonsense knowledge, we created a Question and Answer Dataset with common knowledge of Synonyms (QADS). QADS are questions generated based on SQuAD 2.0 by applying commonsense knowledge of synonyms. The synonyms are extracted from WordNet. Words often have multiple meanings and synonyms. We used an enhanced lesk algorithm to perform word sense disambiguation to identify synonyms for the context.

ELECTRA achieves the state-of-art result on the SQuAD 2.0 dataset in 2019. With about 1/10 scale, ELECTRA can achieve similar performance as BERT does. However, QADS shows that ELECTRA has little ability to handle commonsense knowledge of synonyms. In our experiment, ELECTRA-small can achieve 70% accuracy on SQuAD 2.0, but only 20% on QADS. ELECTRA-large did not perform much better. Its accuracy on SQuAD 2.0 is 88% but dropped significantly to 26% on QADS. In our earlier experiments, BERT, although also failed badly on QADS, was not as bad as ELECTRA. The result shows that even top-performing NLP models have little ability to handle commonsense knowledge which is essential in reading comprehension.

## I. Introduction

ELECTRA [2] is a new method for self-supervised language representation learning. It can be used to pre-train transformer networks using relatively much less compute as compared to other transformers such as BERT. ELECTRA models are trained to distinguish "real" input tokens vs "fake" input tokens generated by another neural network, similar to the discriminator of a GAN. At a small scale, ELECTRA achieves strong results even when trained on a single GPU. At a large scale, ELECTRA achieves state-of-the-art results on the SQuAD 2.0 dataset.

When human beings read, commonsense knowledge is essential and often applied unconsciously. For example, when we read "Which *direction* did Romans use to drift through the Rhine?" or read "Which *way* did Romans use to drift through the Rhine?" We know that the two sentences are of the same meaning because 'direction' and 'way' are synonyms in this context. This type of knowledge is trivial to humans but seems challenging to natural language processing systems.

Constructing an adversarial dataset has been proven to be effective in discovering weak points of machine comprehension models, and in supporting improvement to address the weak points. For example, a major improvement to SQuAD is adding 50,000 questions with no answers. In this research, we developed a Question and Answer Dataset with knowledge of Synonyms (QADS). QADS was automatically generated in the style of SQuAD 2.0 with adversarial questions involving commonsense knowledge of synonyms.

Many words have multiple synonyms and they are often context-dependent. We used Word Sense Disambiguation (WSD) to select the synonyms for a given context. WSD is still a challenging task for computing models. Our algorithm is based on enhanced Lesk algorithms and UKB. UKB is a state-of-the-art system from IXA to form knowledge-based WSD. For the purpose of this study, the enhanced Lesk algorithm has a better balance of performance and robustness. The result set from the WSD processing still contains a large number of synonyms not appropriate for the context. We hired crowd workers to perform the final filtering.

We took SQuAD 2.0 as the baseline in the experiments. ELECTRA-small can achieve 70.01% accuracy on SQuAD 2.0, and ELECTRA-large can achieve 87.85%. This performance leads to the current NLP models. When we test the models on QADS, ELECTRA-small only achieved an accuracy of 20.30%. ELECTRA-large did not perform much better. Its accuracy was only 22.25% on QADS. After cross-verification fine-tune with QADS, the accuracy improved a little bit but was still very low, at about 26%. The result shows that even top-performing NLP models have inadequate ability in handling commonsense knowledge in reading comprehension.

## II. Base Dataset

### A. MRC Dataset

In recent years, many large-scale datasets have been developed, such as SQuAD 1.1 [22], TriviaQA [23], QuAC[21], and COMMENSENQA[16]. These datasets have

*Research supported by NTU-Alibaba JRI Singapore.

Gongqi Lin is with the NTU-Alibaba JRI Singapore (e-mail: lingongqi2009@gmail.com).

Yuan Miao is with the Victoria University, Melbourne Australia (corresponding author to provide phone: 61 3 99194605; e-mail: yuan.miao@vu.edu.au).

Xiaoyong Yang is with the Alibaba, Beijing China (e-mail: xiaoyong.yxy@alibaba-inc.com)

Wenwu Ou is with the Alibaba, Beijing China (e-mail: santong.oww@taobao.com)

Lizhen Cui is with Shandong University, Shandong China (e-mail: clz@sdu.edu.cn)

Wei Guo is with the Shandong University, Shandong China (e-mail: guowei@sdu.edu.cn)

Chunyan Miao is with the Nanyang Technological University, Singapore (e-mail: cymiao@ntu.edu.sg)

greatly sped up the research on Machine Reading Comprehension (MRC). Although many new datasets have been developed based on SQuAD1.1, it is still one of the most notable MRC datasets. SQuAD1.1 involves almost 90,000 training examples and 10,000 validation examples, which was constructed from Wikipedia articles. TriviaQA was designed with question-answer pairs collected retrospectively from Wikipedia and the web. It includes 76,000 training examples and 300 verified validation examples. QuAC is an information-seeking dialog-style dataset. which contains 80,000 training examples and approximately 7,000 validation examples. COMMENSENQA focuses on question answering with commonsense, which requires readers to pick a correct choice based on a given context. Although some progress has been made on the discriminative tasks in MRC, it is still a challenging task for a machine to perform human-like question answering generally. From 2018 to 2019, NLP models started to overtake human beings' performance on some large question and answer datasets, such as SQuAD 1.1.

*B. Adversarial Examples for MRC*

Although NLP models have outperformed human beings on SQuAD1.1, many cases are showing that NLP models are still very weak. Researchers started to develop adversarial question sets to identify the weak points of NLP models. Jia and Liang [24] constructed a set of adversarial examples by adding distractor sentences that do not contradict question answering for humans. Clark and Gardner [25] and Tan et al. [26] used questions to retrieve paragraphs that do not contain the answer and add them as adversarial examples.

Later it has been proven that the adversarial examples generated by rule-based systems are much easier to be detected by large transformers. Thus, Rajpurkar et al. [1] created unanswerable questions by human users and this has been proven to be effective. SQuAD 2.0 dataset [1] added more than 50,000 crowd-sourced unanswerable questions (based on the given articles) into SQuAD v1.1. For SQuAD 1.1, algorithms can assume that for each question, there is an answer, thus they just need to find the most likely answer. While for SQuAD 2.0, many questions do not have an answer. Therefore, when the confidence is low, algorithms must decide between 'no answer' and the most likely answer (although the confidence is low). It does require the algorithms to have a better 'understanding' of the content. Till now, NLP models still fall behind humans on SQuAD 2.0.

*C. Base dataset of QADS*

Because adversarial datasets generated with the rule-based systems are much easier to be handled by nowadays NLP models, we selected SQuAD 2.0 evaluation dataset as the base dataset which already contains 50,000 crowd workers added not answerable questions. An important missing aspect of SQuAD 2.0. is that it did not consider external world knowledge, particularly the commonsense knowledge of synonyms. This is a type of commonsense knowledge we human beings often apply in reading. We used synonyms to replace words in SQuAD 2.0 questions to test how well NLP models can handle the new questions.

We extracted questions and paragraphs from SQuAD 2.0 as the base dataset, noted as the S2E dataset in our process. We then used the NLTK toolkit [29] and spaCy library [30] to process each question in the S2E dataset, filtering out the stop words and numbers, checking the proper nouns and the like to construct a candidate word set. This word set cannot be used for question generation directly. We will need to identify proper synonyms for the context using word sense disambiguation.

III. SYNONYM WORD SENSE DISAMBIGUATION

A word can have multiple synonyms. Some are not appropriate for the given context. We will need to perform word sense disambiguation to select the right synonyms. For example, the house has fourteen meanings in WordNet. Table 1 shows four frequently used meanings, two are nouns and two are verbs.

Depending on the context, human beings can easily select the right meaning of the word. This ability is word sense disambiguation. In the example shown in Figure 1, the selected token ('house') has two target noun synonyms ('social' and 'political'). By checking the context as shown in the text box, a human user can easily tell whether 'an official assembly having legislative powers' is the correct sense or the 'dwelling' one is (exampled in Figure 1).

| Synset | Context | Example |
| --- | --- | --- |
| House.n.01 | A dwelling that serves as living quarters for one or more families | He has a house on Cape Cod. |
| House.n.02 | An official assembly having legislative powers | A bicameral legislature has two houses. |
| House.v.01 | contain or cover | This box houses the gears. |
| House.v.02 | Provide housing for | The immigrants were housed in a new development. |

Table 1 The sample of synset on 'house'

*A. WordNet Concepts*

WordNet [27] is a large lexical English database. In WordNet, nouns, adjectives, verbs, and adverbs are grouped into sets of cognitive synonyms (synsets). Each synset comes with a definition (or gloss) and examples of usage.

Synsets are interlinked with each other through conceptual-semantic and lexical relations. The most frequent encoded relations among synsets are hyponym and hypernym (or ISA relations). A hypernym is a relationship where the word shares meaning with its superordinate words. A hyponym is a relationship where the word generalizes the meaning with its subordinate words. For example, "bed" has a hypernym relationship with "furniture, piece_of_furniture." Conversely, "furniture, piece_of_furniture" has a hyponym relationship with "bed". There is also a similar part-whole relationship between meronym and holonym. "Chair" has a meronym relationship with "back" and "seat." "Bed" has a holonym relationship with "bedroom", "sleeping_room", "sleeping_accommodation", "chamber" and "bed_chamber." A verb synset has entailment relationships such as "buy" to

"pay", or "snore" to "sleep." An adjective synset has antonym relationships such as "wet" to "dry."

For each noun in S2E, we retrieved the corresponding synonyms from WordNet, as well as the corresponding hyponyms, hypernyms, and the corresponding example sentences.

*B. Word Sense Disambiguation*

We employed the enhanced Lesk algorithm (Basile et al., 2014) to select the correct synonyms earlier retrieved from WordNet. The enhanced Lesk algorithm achieved 48.7% accuracy on our dataset. We tried to improve the accuracy by adding the whole paragraph as the context rather than the question only. The accuracy improved by 3%.

We also performed experiments employing UKB [6], which is the state-of-the-art system for knowledge-based WSD. It applies the Personalized Page Rank (PPR) algorithm [7] to an input LKB. However, the result for synonym WSD ranges from 39% to 54% depends on the domains. It only improved the enhanced Lesk algorithm by 2% in the best situation but performed a lot worse in bad situations. Therefore, we chose enhanced Lesk algorithms in most of the cases.

---

**Context**: In November 2006, the Victorian Legislative Council elections were held under a new multi-member proportional representation system. The State of Victoria was divided into eight electorates with each electorate represented by five representatives elected by Single Transferable Vote. The total number of upper house members was reduced from 44 to 40 and their term of office is now the same as the lower house members—four years. Elections for the Victorian Parliament are now fixed and occur in November every four years. Prior to the 2006 election, the Legislative Council consisted of 44 members elected to eight-year terms from 22 two-member electorates.

**Question**: What is the term of office for each house member?

**Selected token**: house

**Synonyms from Wordnet**: 1) A dwelling that serves as living quarters for one or more families; 2) An official assembly having legislative powers.

**Correct contextual synonym**: 2)

---

Figure 1: A Sample of WSD

## IV. DATASET GENERATION

With the synonyms derived with the enhanced Lesk algorithm, we generated the corresponding questions to form the QADS dataset. In this process, we combined two methods: 1) replace only one word with its synonym in the original question, and 2) replace all candidate words with their synonyms.

Given that WSD accuracy rate is not high enough, we appointed crowd workers to verify the generated questions. The crowd workers are native English speakers. All of them either have a bachelor's degree or above, or are enrolled in a bachelor program.

To facilitate the annotation process, we provided each crowd worker unique blocks of questions. Each block contained 2000 ~ 2200 questions from multiple articles. We provided an Excel datasheet generated by the load distribution program for each block so that crowd workers could perform the verification and log the progress. Crowd workers were asked to check whether the provided synonyms for the original token are correct or not. Crowd workers could add synonyms based on their commonsense knowledge.

As a further measure to ensure the quality, the crowd workers were also asked to annotate the following cases

(i) the original SQuAD question was wrong (meaningless by itself),

(ii) idiomatic expressions (not a synonym of a single word but the whole idiomatic phrase),

(iii) fixed prepositional phrase, such as 'according to', 'in terms of', 'in favor of', etc, (not synonym of a single word but the whole phrase),

(iv) named entities consisting of multiple words, and

(v) the original SQuAD question was an improper sentence but could be fixed by supplying the missing verbs or other keywords; (about 5% of SQuAD questions are improper sentences).

To guarantee the quality, crowd workers were rewarded only if their verifications met the minimum requirements. They needed to ensure 90% of their verification were properly completed. We randomly sampled 15% ~ 20% of crowd workers' work to double-check.

Some of the articles and the corresponding questions are domain-specific. The corresponding commonsense knowledge of synonyms are therefore domain specific as well. To ensure that the crowd worker has the domain-specific commonsense knowledge, we developed an algorithm to automatically identify the domain for each dataset. This algorithm

- randomly sampled 100 questions from each question set,
- extracted the domain of each word in the 100 questions using the WordNet domain function [29],
- ranked all the possible domains by their occurrence,
- the top two domains were selected as the domain of the question set.

The crowd workers appointed for the question set were required to have the corresponding domain knowledge.

## V. EXPERIMENTS

Our goal is to test the latest NLP models' ability in applying commonsense knowledge of synonym as what human readers do. If there is a gap, the dataset (QADS) can provide researchers insights on possible improvement methods. We selected the state-of-the-art NLP model ELECTRA. ELECTRA's performance was announced in late 2019 and the model was recently released.

*A. Pre-Trained Models*

Pre-training is crucial for machine reading comprehension since models learn the contextual representation with a semantic relation among words. To address the context-dependent nature of words, we need to distinguish the semantics of words in different contexts by using contextual embeddings. In practice, a fully-connected graph is applied to model the relation of two words, and the model will learn the structure by itself. However, the number of model parameters increases rapidly while adding deeper layers, which consumes significant computing resources and time to train a model. Additionally, overfitting and building large-scale labeled datasets are very challenging due to the extremely expensive resource and effort(time) requirements.

On the other hand, we can easily obtain the large-scale unlabeled corpora, and we can learn a good representation from them by developing deep learning algorithms, such as BERT [4], GPT [11], etc. As an approach to transfer learning, we can use these representations to fine-tune on down-stream tasks, such as language modeling, question answering, text classification, etc. Large pre-trained models [11, 12, 14, 15, 18] have brought dramatic empirical improvements on these tasks in the past few years. Most of these pre-trained models are based on Transformers [20]. The Transformer is a model that uses multiple self-attention layers to boost the train speed. The terminology "self-supervised" means that supervisions used for pre-training are automatically collected from the raw data without manual annotation. Dominant learning objectives are language modeling and its variations. For example, GPT-2 [28] was trained to predict the next word in 40GB of Internet text. GPT-2 is a large transformer-based language model with large parameters, which displays the ability to do language tasks, such as reading comprehension, question answering, translation, summarizations, using no task-specific training data. But the ultimate goal of pre-training is not to just train a good language model but to consider both preceding and following contexts to learn general-purpose contextual representations. For example, BERT [4] learns to predict the masked word of a randomly masked word sequence given surrounding contexts.

*B. ELECTRA*

ELECTRA [2] utilizes the Replaced Token Detection (RTD) with a generator to replace some tokens of a sequence. There are two stages to train a generator and a discriminator, which are specified as follows: (1) the generator is trained with the MLM task by fixed steps; (2) the discriminator is initialized with the weights of the generator. After these two stages, the discriminator is trained with a discriminative task. At this stage, the generator keeps frozen. Here the discriminative task indicates justifying whether the input token has been replaced by the generator or not. The generator is discarded after pre-training, and only the discriminator will be fine-tuned on downstream tasks.

ELECTRA includes three different-sized models: small, base, and large. ELECTRA-Small is to improve the efficiency of pre-training. It is a model that can be quickly trained on a single GPU. This work significantly reduces the BERT-Base hyperparameters: reducing sequence length from 512 to 128, reducing the batch size from 256 to 128, etc. Despite its much smaller size, the performance of ELECTRA-Small on most NLP tasks can achieve a similar level of the state-of-the-art MRC models, such as BERT, ELMo, and GPT. ELECTRA-Base yields better results than BERT-Base and XLNet-Base, and even surpassing BERT-Large according to most metrics.

The ELECTRA-Large models are of the same size as BERT-Large, thus the training needs much longer time as compared to the base model. This model uses a batch size 2048 and XLNet pre-training data. ELECTRA-Large model outperforms RoBERTa and BERT, which achieved 88.0% for Exact Match (EM) on SQuAD 2.0 Dev dataset, which also outperformed BERT WWM.

For SQuAD, ELECTRA added the question answering module from XLNet. It is slightly more sophisticated than BERT because it jointly rather than independently predicts the start and end positions. It also added an "answerability" classifier for SQuAD 2.0.

*C. Environmental Setting*

The MRC dataset used in this paper is SQuAD 2.0, which combines SQuAD 1.1 (contains over 100,000 passage-question pairs) with over 50,000 unanswerable questions. The dataset has been randomly partitioned into three parts: a training set (80%), a development set (10%), and a test set (10%).

We mainly worked on the development set (35 articles) to generate the new question set QADS. We tokenized the MRC dataset with spaCy 2.0.13, using enhanced Lesk algorithm and WordNet 3.0 with NLTK 3.3, and developed QADS with Tensorflow-GPU 2.0.0 in an Anaconda environment.

We evaluated pre-trained ELECTRA [2] implemented in Tensorflow-TPU.

- ELECTRA-Small: 12-layer, 256-hidden, 110M parameters
- ELECTRA-Base: 12-layer, 768-hidden, 110M parameters
- ELECTRA-LARGE: 24-layer, 1024-hidden, 335M parameters

For find-tuning, we followed the same way that the authors of ELECTRA published on github (using TPU). Taking ELECTRA-base model as an example, we used the following command to fine-tune with QADS:

python3 run_finetuning.py --data-dir $DATA_DIR --model-name electra_base --hparams '{"model_size": "base", "task_names": ["squad"], "user_tpu": "True",

"tpu_name": "grpc://tpu_address", "num_tpu_cores": 8}'

*D. Model Evaluation*

We evaluated the models on the QADS dataset in two ways.

- QADS only: Fine-tuned on SQuAD 2.0 and evaluated on QADS.
- SQuAD 2.0 + QADS: Fine-tuned on SQuAD 2.0 first, then fine-tuned again using cross-validation on QADS. The cross-validation fine-tuning followed the standard cross-validation setting: the training data are 80% of QADS, and evaluation was performed on the remaining 20% of QADS.

*E. Results*

The results of the experiments of the above configurations are shown in Table 2. As a baseline, we used the pre-trained models based on the full SQuAD 2.0 training sets and evaluate them on SQuAD 2.0 development sets. The resulting accuracy is 87.85% by ELECTRA-Large and 83.27% by ELECTRA-Base. To compare the results, we used the pre-trained models to evaluate them on the QADS dataset. This produced an accuracy of 22.15% by ELECTRA-Base and 22.25% accuracy by ELECTRA-Large.

Cross-validation fine-tuning on QADS helped ELECTRA but not much. The ELECTRA-Large achieved only 25.93%. Although expected, the result is surprisingly low, showing some essential mechanism was missing in the model in handling commonsense knowledge of synonyms.

| Dataset | ELECTRA-Small (%) | ELECTRA-Base (%) | ELECTRA-Large (%) |
|---|---|---|---|
| SQuAD 2.0 | 70.1 | 83.27 | 87.85 |
| QADS | 20.3 | 22.15 | 22.25 |
| SQuAD 2.0 + QADS | 21.32 | 25.64 | 25.93 |

Table 2 Experiment results

*F. Results Discussion*

The results show that ELECTRA's performance on QADS is substantially lower than that on SQuAD2.0. The ELECTRA-base model only achieved 22.15% accuracy. Even the ELECTRA-Large model did not improve much. It achieved only 22.25% on the QADS dataset. Then we considered SQuAD 2.0 + QADS solution, which fine-tuned on SQuAD2.0 dataset firstly, and fine-tuned on the cross-validation QADS dataset. The results are shown in Table 2, and ELECTRA-Large still only achieved 25.93% accuracy on the evaluation set.

The results proved that QADS forms a challenging extension to SQuAD 2.0. It revealed that ELECTRA has little ability in handling commonsense knowledge of synonyms. Even when we trained it on SQuAD 2.0, and then fine-tuned on QADS, the system still failed badly in answering QADS questions, which human readers see little difference between the QADS and SQuAD 2.0 questions.

The results suggested two future works for dataset construction. The first one is, rather than using a single rule (e.g. synonyms), to generate questions from one monolithic dataset (e.g. SQuAD 2.0), the community may be able to construct more questions by aggregating a variety of distinct, independent datasets following a similar format. For example, pre-training on SQuAD and evaluating QADS proves to be more challenging than training and testing on SQuAD alone. Secondly, if we wish to use the approach to automatically generate more questions to grow QADS, which currently has 5000+ questions, the WSD needs to be significantly improved to minimize the needs of crowd workers.

Interestingly, although ELECTRA is an improved transformer originated from BERT, BERT's performance on commonsense knowledge is a lot better. Its exact match rate dropped to 63.6% [31]. Our work found that the 63.6% correct answers contained about 20% 'lucky hit', which we can identify and remove. The result will be reported soon.

## VI. CONCLUSION

In this paper, we showed that the state-of-the-art NLP model, ELECTRA, has failed badly on questions involving commonsense knowledge of synonyms. ELECTRA has outperformed BERT on SQuAD 2.0 dataset, achieved an accuracy of 87.85%. However, on QADS, its accuracy dropped to 25.93%, even worse than BERT. To human users, QADS questions read almost the same as that in SQuAD 2.0, only some words have been replaced with the synonyms.

To generate QADS, we applied the enhanced Lesk algorithm to perform word sense disambiguation for the synonyms retrieved from WordNet. The WSD was not accurate enough thus we appointed crowd workers to verify the autogenerated questions.

The significant performance gap shows that QADS questions present a challenging extension of the SQuAD 2.0 dataset, which will help researchers to develop better NLP models, particularly in reading with commonsense knowledge as what human readers do.

ACKNOWLEDGMENT

This research has been supported by NTU-Alibaba Joint Research Institute, Singapore.

We thank Lutfar Khan for his valuable help and comments on the paper.